# Bayesian Network Classifiers in a High Dimensional Framework

**Tatjana Pavlenko**
Dept. of Engineering, Physics and Mathematics
Mid Sweden University
851 70 Sundsvall
Sweden
tatjana@fmi.mh.se

**Dietrich von Rosen**
Dept. of Biometry and Informatics
Swedish University of Agricultural Sciences
Box 480, SE-750 07 Uppsala
Sweden
Dietrich.von.Rosen@bi.slu.se

## Abstract

We present a *growing dimension asymptotic* formalism. The perspective in this paper is classification theory and we show that it can accommodate probabilistic networks classifiers, including naive Bayes model and its augmented version. When represented as a Bayesian network these classifiers have an important advantage: The corresponding discriminant function turns out to be a specialized case of a *generalized additive model*, which makes it possible to get closed form expressions for the asymptotic misclassification probabilities used here as a measure of classification accuracy. Moreover, in this paper we propose a new quantity for assessing the discriminative power of a set of features which is then used to elaborate the augmented naive Bayes classifier. The result is a weighted form of the augmented naive Bayes that distributes weights among the sets of features according to their discriminative power. We derive the asymptotic distribution of the sample based discriminative power and show that it is seriously overestimated in a high dimensional case. We then apply this result to find the optimal, in a sense of minimum misclassification probability, type of weighting.

## 1 INTRODUCTION

*Bayesian networks* (also referred to as *probabilistic networks* or *belief networks*) have been used in many areas as convenient tools for presenting dependence structures. A few examples are [2,8,17], also including artificial intelligence research, where these models have by now established their position as valuable representations of uncertainty.

Bayesian network models offer complementary advantages in classification theory and pattern recognition tasks such as the ability to deal effectively with uncertain and high dimensional examples. Several approaches have recently been proposed for combining the network models and classification (discrimination) methods; see for instance [10,12]. In the current paper we also combine the network approach with discrimination and provide some new ideas which are of principal interest.

The focus of this paper is on the classification task, where Bayesian networks are designed and trained essentially to answer the question of accurate classifying yet unseen examples. Assuming classes to be modeled by a probability distribution, quantifying a Bayesian network amounts to assessing these distributions (densities) to the network's variables conditional on their direct predecessors in the graph. The problem of assessing a distribution in statistics as well as in pattern recognition can be viewed as a problem of estimating a set of parameters associated with a specific parametric family of distributions that is given a priori. However, applying parametric estimation techniques naturally raises the following question: What can we say about the accuracy of the induced classifier, *given* that the network structure is known or assumed to be correct and we are training on a possibly small sample of observations? Clearly, in such a case classification will be affected by inaccuracy of involved estimates; the resulting classifier will especially degrade in performance in a high dimensional setting, i.e. when the size of the training data set is comparable to the dimensionality of observations. The degradation effect is known as "curse of dimensionality", a phrase due to Bellman, [1], and one of our primary goal is to explore this effect when using Bayesian network classifiers.

In order to tackle this problem effectively, we employ a *growing dimension asymptotic* formalism [15], essentially designed for multivariate problems and show how it can accommodate such classifiers as naive Bayes [3] and its augmented version [6]. The technique we develop in this study exploits the main assumption encoding in these networks, namely that each feature variable (set of the features in augmented naive Bayes) is independent from the rest of features given the root of the network (viewed in the present setting as a class variable). This provides a factorized representation of class distributions and we show that the corresponding discriminant for both types of the network turns out to be a specialized case of the *generalized additive models* [9]. The main advantage here is that given the additive structure of the discriminant, the Gaussian approximation can be applied to its distribution in a high dimensional setting. As we show in the present study, this in turn yields explicit expressions for the misclassification probabilities which are used as a measure of classification accuracy.

In addition, we introduce a new technique for measuring discriminative power of a given set of features which is based on the cross-entropy distance between the classes. This measure is then used to extend the augmented naive Bayes classifier to deal with sets of features with a priori different discriminative power. The extended classifier in this approach is naturally a *weighted* version of the augmented naive Bayes. The analysis of performance of this modified classifier in a high dimensional setting, and the optimal in a sense of minimum misclassification probability choice of weighting are the further subjects of the current study.

## 2 OVERVIEW OF BAYESIAN NETWORK CLASSIFIERS

In the classification problem, the goal is to build an accurate classifier from a given data set $(\mathbf{x}, \mathcal{C})$, where each $\mathbf{x}$ consists of $p$ feature variables $x_1, \ldots, x_p$ together with a value for a class variable $\mathcal{C}$. In the sequel we assume the feature variables to be continuous, which makes it possible to use various parametric models for the joint probability density $f(\mathbf{x}, \mathcal{C}; \theta)$ of a random sample $(\mathbf{x}, \mathcal{C})$, where $\theta \in \Theta$ is the vector of (unknown) parameters needed to describe the $j$th class density, $j = 1, \ldots, \nu$.

In the probabilistic framework, the main goal is to evaluate the classification posterior distribution $\Pr(\mathcal{C} = j | \mathbf{x}; \theta)$. We can further distinguish two different approaches for estimating this distribution: in the *diagnostic paradigm* one tries to model $\Pr(\mathcal{C} | \mathbf{x}; \theta)$ directly, while in the *sampling paradigm* the interest centers on $f^j(x; \theta)$, and we have the following factorization $f(\mathbf{x}, \mathcal{C}; \theta) = \pi_j f^j(\mathbf{x}; \theta)$, with the prior probabilities $\pi_j$ for the classes assumed to be known or estimated from the proportions in the training set, i.e. $\hat{\pi}_j = n_j / \sum n_k$. In this study we will concentrate on the sampling paradigm, which means that we assume a stationary underlying joint distribution (density) $f^j(\mathbf{x}) = f^j(\mathbf{x}; \theta)$ over the $p$ feature variables $x_1, \ldots, x_p$. The desired posterior probability can then be computed by using the Bayes' formula, which tells us that $\Pr(\mathcal{C} = j | \mathbf{x}; \theta) \propto f^j(\mathbf{x}; \theta) \pi_j$. It is relatively simple to show that the rule that classifies to the largest posterior probability will give the smallest expected misclassification rate. This rule is known as *Bayes' classifier*.

As a model family for the underlying class distribution we consider in this study a finite number of Bayesian network models. As a quick synopsis: A Bayesian network [17] is the graphical representation of a set of independence assumptions holding among the feature variables. These independencies are encoded in a directed acyclic graph, where nodes correspond to the features $x_1, \ldots, x_p$ and the arcs represent the direct probabilistic influence between its incident nodes. Absence of an arc between two variables means that these variables do not influence each other directly, and hence are (conditionally) independent. In this way a Bayesian network specifies a complete joint probability distribution over all feature variables which then can be used for producing classifiers.

A well-known example of the sampling Bayesian network classifier is the *naive Bayesian classifier* [3,6] (or simply naive Bayes), which is a random network with one arc from the class node to each of the feature nodes. This graph structure represents the assumption that the feature variables are conditionally independent of each other, given the class, from which it is immediate that $f^j(\mathbf{x}; \theta) = \prod_{i=1}^{p} f_i^j(\mathbf{x}_i; \theta)$, $j = 1, \ldots, \nu$. Hence, using Bayes' formula we get $\Pr(\mathcal{C} = j | \mathbf{x}) \propto \pi_j \prod_{i=1}^{p} f_i^j(\mathbf{x}_i; \theta)$. It is worth noting that the naive Bayes does surprisingly well when only a finite sample of training observation is available; This behavior has been noted in [14]. The naive Bayes approach also turns out to be effective when studying the curse of dimensionality effect; see for instance [4], where the bias and variance induced on the class density estimation by the naive Bayes decomposition and their effect on classification have been studied in a high dimensional setting.

However, the structural simplicity of the naive Bayes classifier calls for developing better alternatives. In [6] the problem was approached by *augmenting* the naive Bayes model by allowing additional arcs between the features that capture possible dependencies among them. This approach is called *augmented naive Bayes* and approximates interactions between features using a tree-structure imposed on the naive Bayes structure. Classification accuracy has further been studied in [7], where the augmented classifier has been extended to directly model the distributions of continuous feature variables by parametric and semi parametric models.

Intuitively, one would expect the emerging feature variables in this framework to be such that the features in the same subset are highly dependent on each other, so that the relevant feature subset form a partitioning of the features. Various techniques of finding such subsets and their optimal size are proposed and experimentally studied in [12]. In the sequel, we focus on the augmented naive Bayes model where the partitioning is induced to form $\kappa$ $m$-dimensional *blocks* of features or subnetworks so that $p = m\kappa$, and

$$\underbrace{x_{11},...,x_{1m}}_{\mathbf{x}_1},\underbrace{x_{21}..,x_{2m}}_{\mathbf{x}_2},...,\underbrace{x_{\kappa 1},..,x_{\kappa m}}_{\mathbf{x}_\kappa},$$
$$\underbrace{\theta_{11},...,\theta_{1m}}_{\theta_1},\underbrace{\theta_{21}..,\theta_{2m}}_{\theta_2},...,\underbrace{\theta_{\kappa 1},..,\theta_{\kappa m}}_{\theta_\kappa}, \quad (1)$$

where $i$th subnetwork consists of $m$ nodes $x_{i1},...,x_{im}$ and is fully connected, but there are no arcs between the blocks. Observe that we require the resulting blocks to be disjoint and non-empty independent (no arcs between the blocks) $m$-dimensional subsets and assume for convenience the same decomposition of $\theta$, i.e. the dimension $m$ of each $\theta_i^j$ is identical throughout of all blocks. Actually, as it will be observed later, for our purposes it is enough to assume that the block size is *fixed*, to some constant which could be different for different blocks. Clearly, with the partitioning just described, the number of block is bounded from above by the total number of features.

In the following we will focus on binary classification, the special (but common) case in which $\nu = 2$. Although most of the concepts generalize to the case $\nu \geq 3$, the derivations and underlying intuition are more straightforward for this special "two-class" case. Friedman (see [5]) suggested the following generalization for the multi-class setting: Solve each of the two-class problems, and then for a test observation, combine all the pairwise decisions to form a $\nu$-class decision. Observe that Friedman's combination rule is quite intuitive: Assign to the class that wins the most pairwise comparisons. For convenience in what follows, we will make use of the decision boundaries that are expressed in terms of a *discriminant function*,

$$\mathcal{D}(\mathbf{x};\theta^1,\theta^2) = \ln \frac{f^1(\mathbf{x};\theta^1)}{f^2(\mathbf{x};\theta^2)}. \quad (2)$$

To motivate why this representation of the classifier is attractive, note the discriminant preserves the ordering of the class posterior probabilities leading to the decision rule:

$$\mathcal{C}(\mathbf{x}) = \begin{cases} 1 & \text{whenever} \quad \mathcal{D}(\mathbf{x};\theta^1,\theta^2) > \ln \frac{\pi_2}{\pi_1} \\ 2 & \text{otherwise} \end{cases} \quad (3)$$

The rule is simple to describe and implement and can consequently be used to classify yet unseen examples. Another advantage of using the discriminative formulation is that the performance accuracy of $\mathcal{D}(\mathbf{x};\theta^1,\theta^2)$ can be measured by misclassification probabilities defined as follows:

$$\mathcal{E}_1 = \Pr(\mathcal{D}(\mathbf{x};\theta^1,\theta^2) \leq \ln \frac{\pi_2}{\pi_1}|\mathcal{C}(\mathbf{x}) = 1),$$
$$\mathcal{E}_2 = \Pr(\mathcal{D}(\mathbf{x};\theta^1,\theta^2) > \ln \frac{\pi_2}{\pi_1}|\mathcal{C}(\mathbf{x}) = 2). \quad (4)$$

These can then form the *Bayes risk* $\mathcal{R}_{\mathcal{D}(\mathbf{x};\theta^1,\theta^2)} = \pi_1\mathcal{E}_1 + \pi_2\mathcal{E}_2$, which in turn gives a straitforward way of judging the classification accuracy. Note also that in the symmetric case with equal prior probabilities both class-wise error rates are equal, and the minimum attainable by Bayes risk is $\mathcal{R}_{\mathcal{D}(\mathbf{x};\theta^1,\theta^2)} = \frac{1}{2}(\mathcal{E}_1 + \mathcal{E}_2)$.

Observe that given the network structure, learning the classifier $\mathcal{D}(\mathbf{x};\theta^1,\theta^2)$ reduces to filling in the parameters by computing the appropriate values of $\hat{\theta}^j$ from the training set of data (the detailed properties of $\hat{\theta}^j$ are described in the next section). This so-called *plug-in approach* while very simple and asymptotically correct given that dimensionality $p$ is fixed (as the sample size increases, the plug-in density estimate $\hat{f}^j(\mathbf{x};\hat{\theta}^j)$ converges to the underlying one), yields highly biased estimates in a case of too many feature variables, such as pixels of a digitized image, for example. This naturally hurts the corresponding classification procedure and it will not likely achieve minimum misclassification probability, even asymptotically. Our goal is to show to what extent classification accuracy of Bayesian network classifiers suffers if the number of observations are few relative to their dimensionality (high-dimensional setting). In the next sections, we describe a technique that attends to this issue by generalizing the relationship between the number of features (blocks) and sample size and exploiting independences induced by the network structure.

## 3　HIGH-DIMENSIONAL FRAMEWORK

We begin by introducing the *growing dimension asymptotics* and then show how this formalism can accommodate Bayesian network classifiers, including naive Bayes and its augmented version.

A theoretically sound way to deal with the high-dimensional problem is to turn to a general asymptotic approach, meaning that a relationship between dimensionality and sample size satisfies the condition: $\lim_{n_j \to \infty} \lambda(p, n_j) < \infty$, where $\lambda(p, n_j)$ is a positive function increasing along $p$ and decreasing along $n_j$, $j = 1,2$. Since the increase of $p$ and $n_j$ is somehow simultaneous in a high-dimensional setting, the asymp-totic approach we are going to work with can be based on the ratio

$$\lim_{n_j \to \infty} \frac{p}{n_j} = c, \qquad (5)$$

where $0 < c < \infty$ is a certain constant for each $j = 1, 2$. This approach is often referred to under the name of *growing dimension asymptotics* and our goal is to apply this to explore the curse of dimensionality effect on the classification accuracy. Regarding $n_j$, in this study we assume the same rate of growing for both samples sizes so that $n_1 = n_2 = n$.

In order to completely specify the learning method in the context of augmented naive Bayes model, we define the asymptotic properties of estimates $\hat{\theta}_i^j$ to be plugged-in into $\mathcal{D}(\mathbf{x}; \theta^1, \theta^2)$. We introduce the statistics $T_i^j = n^{1/2}(\hat{\theta}_i^j - \theta_i^j)' I^{1/2}(\theta_i^j)$, which for each $i = 1, \ldots, \kappa$ describes the standardized bias of the estimate $\hat{\theta}_i^j$, where $I^j = I(\theta^j)$ is the Fisher information matrix which is positive definite for all $\theta^j \in \Theta^j$ and whose eigenvalues are bounded from above. By the network structure, the matrices are of block-diagonal form with blocks $I_i^j = I(\theta_i^j)$ of dimension $m \times m, j = 1, 2$. We assume that the estimate $\hat{\theta}_i^j$ is such that for each $j$ uniformly in $i$:

1. $\lim_{n \to \infty} \max_i |\mathsf{E}[T_i^j]| = 0$.

2. All eigenvalues of the matrices $n\mathsf{E}[(\hat{\theta}_i^j - \theta_i^j)(\hat{\theta}_i^j - \theta_i^j)']$ are bounded from above so that

$$\lim_{n \to \infty} \max_i |n\mathsf{E}[(\hat{\theta}_i^j - \theta_i^j)' I(\theta_i^j)(\hat{\theta}_i^\nu - \theta_i^j)] - m|$$
$$= \lim_{n \to \infty} \max_i |\mathsf{E}[\langle T_i^j, T_i^j \rangle] - m| = 0, \qquad (6)$$

where $\langle \bullet, \bullet \rangle$ denotes the scalar product.

3. $\max_i \mathsf{E}[|T_i^j|^3] = \mathcal{O}(\frac{1}{n^{3/2}})$.

4. The asymptotic distribution of $T_i^j$ converges to $\mathcal{N}_m(0, I)$ as $n$ approaches infinity.

These assumptions form the standard set of "good" asymptotic properties, of which first three reflect unbiasedness, efficiency and boundness of the third absolute moment of $\hat{\theta}_i^j$, uniformly in $i$ as $n \to \infty$.

Let us now in this framework have a look at the structure of the plug-in discriminant $\mathcal{D}(x; \hat{\theta}^1, \hat{\theta}^2)$, given partitioning of features induced by augmented naive Bayes. Since we fix the size of the relevant blocks to the constant $m$, the total number of blocks $\kappa$, must grow together with $n$ according to (5) in such a way that

$$\lim_{n \to \infty} \frac{\kappa}{n} = \rho, \quad \text{where} \quad 0 < \rho < \infty \qquad (7)$$

and $c = m\rho$. This assumption being designed for the special dependence structure among the features, is just a particular case of (5).

Furthermore, using the idea of augmenting and block independence, we can decompose the discriminant according to the structure of the network so that

$$\mathcal{D}(\mathbf{x}; \hat{\theta}^1, \hat{\theta}^2) = \sum_{i=1}^{\kappa} \mathcal{D}_i(\mathbf{x}_i; \hat{\theta}_i^1, \hat{\theta}_i^2) \qquad (8)$$

where $\mathcal{D}_i(\mathbf{x}_i; \hat{\theta}_i^1, \hat{\theta}_i^2) = \ell_i(\mathbf{x}_i; \hat{\theta}_i^1) - \ell_i(\mathbf{x}_i; \hat{\theta}_i^2)$ and $\ell_i(\mathbf{x}_i, \theta_i^j) := \ln f_i^j(\mathbf{x}_i, \theta_i^j)$. This implies that the discriminant induced by augmented naive Bayes network turns out to be *log additive* in each block of features and the corresponding discriminative procedure is a special case of the *Generalized Additive Models*; see [9]. It is important to note that with this representation, the naive Bayes model can be viewed as a particular case of the augmented one: Indeed, if we assume that $m = 1$, (and so $\kappa = p$), then the resulting discriminant $\mathcal{D}(\mathbf{x}; \hat{\theta}^1, \hat{\theta}^2) = \sum_{i=1}^{p} \mathcal{D}_i(\mathbf{x}_i; \hat{\theta}_i^1, \hat{\theta}_i^2)$ that corresponds to the usual naive Bayes network.

The main advantage of the additive structure of the discriminant is that in the asymptotic framework described above, $\mathcal{D}(\mathbf{x}; \hat{\theta}^1, \hat{\theta}^2)$ can be viewed as a sum of a growing numbers of independent random variables (number of blocks, $\kappa$ grows together with $n$). Hence, one could expect that the central limit theorem is applicable to this sum. This approach has been studied in detail in [15], where we have estimated the first three moments of $\mathcal{D}(\mathbf{x}; \hat{\theta}^1, \hat{\theta}^2)$ and then applied the Liapunov theorem, which states the convergence of the sum towards a Gaussian distribution. We have also proved that the first two moments can be obtained in terms of the cross-entropy distance between the classes, block size $m$ and high dimensionality factor $\rho$. The cross-entropy distance, written as

$$\mathcal{J} = \int \ln \frac{f^1(\mathbf{x}; \theta^1)}{f^2(\mathbf{x}; \theta^2)} \Big( f^1(\mathbf{x}; \theta^1) - f^2(\mathbf{x}; \theta^2) \Big) d\mathbf{x}$$

is defined as a symmetric combination of the Kullback-Leibler divergences between the class distributions $f^1(\mathbf{x}, \theta^1)$ and $f^2(\mathbf{x}, \theta^2)$; see for instance, [13]. Observe that by the network structure the distance $\mathcal{J} := \mathcal{J}(\kappa)$ is decomposable as

$$\mathcal{J}(\kappa) = \sum_{i=1}^{\kappa} \mathcal{J}_i, \qquad (9)$$

where $\mathcal{J}_i = \int \ln \frac{f_i^1(\mathbf{x}_i, \theta_i^1)}{f_i^2(\mathbf{x}_i, \theta_i^2)} (f^1(\mathbf{x}, \theta^1) - f^2(\mathbf{x}, \theta^2)) d\mathbf{x}$ is the input of $i$th block into the distance $\mathcal{J}(\kappa)$. In the asymptotic framework it is worthwhile introducing a distribution function of the block distances

$$H_\kappa(\gamma^2) = \frac{1}{\kappa} \sum_{i=1}^{\kappa} \mathbf{1}_{\{\frac{n \mathcal{J}_i}{2}, \infty\}}(\gamma^2),$$

where $\mathbf{1}_A$ is the indicator function of the set $A$. We suppose also that the convergence $\lim_{\kappa \to \infty} H_\kappa(\gamma^2) = H(\gamma^2)$ takes place uniformly in $\gamma^2$ and $H(\gamma^2)$ is a known distribution. The limiting value $\mathcal{J}$ of the distance $\mathcal{J}(\kappa)$ given the distribution $H(\gamma^2)$ is then naturally defined by replacing the sum in (9) with an integral equals $\mathcal{J} = 2 \int \gamma^2 dH(\gamma^2)$. Using these results and adding supplementary regularity conditions of Cramér type on the functions $\ell_i(\mathbf{x}_i; \theta_i^j)$, we find that

$$\mathsf{E}[\mathcal{D}(\mathbf{x}; \hat{\theta}^1, \hat{\theta}^2)] \to \mathcal{J}/2, \mathsf{Var}[\mathcal{D}(\mathbf{x}; \hat{\theta}^1, \hat{\theta}^2)] \to \mathcal{J} + 2m\rho$$

as $n \to \infty$. The third moment $\mathsf{E}[\mathcal{D}(\mathbf{x}, \hat{\theta}^1, \hat{\theta}^2)]^3$ is of the order $\mathcal{O}(n^{-1/2})$ which implies that the conditions of the Liapunov theorem are satisfied. Taking into account this assertion and using the expressions for the misclassification probabilities given by (4) produces the following

**Theorem 1:** *Assuming that $\mathcal{D}(\mathbf{x}; \hat{\theta}^1, \hat{\theta}^2)$, induced by the augmented naive Bayesian network is decomposable according to (8) and the estimates $\hat{\theta}_i^j$ satisfy the set of asymptotic conditions (6) uniformly in $i$, the misclassification probabilities have the limits*

$$\mathcal{E}_1 \to \Phi\left(-\frac{\mathcal{J} - \pi_0}{2\sqrt{\mathcal{J} + 2m\rho}}\right), \mathcal{E}_2 \to \Phi\left(-\frac{\mathcal{J} + \pi_0}{2\sqrt{\mathcal{J} + 2m\rho}}\right)$$

as $n \to \infty$, where $\Phi(y) = \frac{1}{\sqrt{2\pi}} \int_{-\infty}^y \exp(-z^2/2) dz$ and $\pi_0 = \ln \frac{\pi_2}{\pi_1}$.

These closed form expressions for $\mathcal{E}_j$ allow us to further study the classification accuracy and highlight the curse of dimensionality effect. Observe that the limiting value of $\mathcal{R}_{\mathcal{D}(\mathbf{x}; \hat{\theta}^1, \hat{\theta}^2)}$ achieves a minimum when $\mathcal{E}_1 = \mathcal{E}_2$ and $\pi_1 = \pi_2$, so that

$$\mathcal{R}_{\mathcal{D}(\mathbf{x}; \hat{\theta}^1, \hat{\theta}^2)} \longrightarrow \Phi\left(-\frac{\sqrt{\mathcal{J}}}{2} \frac{1}{\sqrt{1 + \frac{2m\rho}{\mathcal{J}}}}\right),$$

as $n \to \infty$. In order to understand the effect, this result has to be compared with the minimal misclassification risk for ideal case of exactly known class densities: $\mathcal{R}_{\mathcal{D}(\mathbf{x}; \theta^1, \theta^2)} = \Phi\left(-\frac{\sqrt{\mathcal{J}}}{2}\right)$ (for this result, see for instance [16]). As can be seen, the term $\frac{2m\rho}{\mathcal{J}}$ is negligible in the standard asymptotic setup, i.e when the number of unknown parameter is fixed and the sample size is sufficiently large. In this case $\rho = 0$ and the two discriminants (with and without estimation) turns out to be asymptotically equivalent. However, more realistic is the situation reflected by the growing dimension asymptotics, i.e. when the number of unknown parameters is comparable to the sample size. In this case, filling the unknown parameters with their sample analogous leads to the increase of misclassification risk, governing by $\frac{2m\rho}{\mathcal{J}}$.

## 4 WEIGHTING THE AUGMENTED NAIVE BAYES CLASSIFIER

While the problem of dimensionality when using Bayesian network classifiers has received some attention in the recent literature, a measure of features (or sets of features) discriminative power reflecting their ability to distinguish classes has especially been overlooked in the development of learning Bayesian networks. In this section, we show how to estimate and take into account the discriminative power of a set of features in the context of the augmented naive Bayes classifier. Of course, what we need to deal with this problem, is a suitable (for the classification task) measure of the discriminative power of the blocks. Since we measure performance accuracy by the misclassification probability, the latter seems to be most appealing function for this. However, in this study we propose a *distance-based* technique which exploits the following relationship: misclassification probability is a monotone decreasing function of the distance between classes (see, for instance the results from the previous section) which means that distance-based measure, defined by $\mathcal{J}_i$, $i = 1, \ldots, \kappa$ induces, over the set of all potential blocks, the same ranking as that one induced by $\mathcal{R}$.

Our approach is as follows. Features discriminative power is incorporated into the augmented discriminant by means of a *weight function*, defined by $w_i := w(\frac{n\hat{\mathcal{J}}_i}{2})$ for the $i$th set of features, $i = 1, \ldots, \kappa$ where $w_i(u)$ is nonnegative and bounded for $u > 0$. A modified discriminant can then be expressed as

$$\mathcal{D}_w(\mathbf{x}, \hat{\theta}^1, \hat{\theta}^2) = \sum_{i=1}^\kappa w_i \mathcal{D}_i(\mathbf{x}_i, \hat{\theta}_i^1, \hat{\theta}_i^2), \qquad (10)$$

where each $\mathcal{D}_i(\mathbf{x}_i, \hat{\theta}_i^1, \hat{\theta}_i^2)$ is weighted according to the sample based discriminative power $\frac{n\hat{\mathcal{J}}_i}{2}$ of the $i$ block of features. The novelty of this technique is that weighting by means of discriminative power counteracts equalizing the inputs of the low- and high- relevant blocks, inherent in the augmented naive Bayes model. Observe however, that with the latter technique, the true discriminative power can be severely affected by the curse of dimensionality. To give an impression about this effect, we find here the asymptotic distribution of the estimator $\frac{n\hat{\mathcal{J}}_i}{2}$ and derive the bias induced by sample based weighting scheme (10).

**Theorem 2:** *Let $\chi(u; m, \gamma^2)$ be the density function of a non-central $\chi^2$ random variable with $m$ degrees of freedom and non-centrality parameter $\gamma^2$. Let also $g(u; \gamma_i^2)$ be a density function of $\frac{n\hat{\mathcal{J}}_i(n)}{2}$, where*

$$\gamma_i^2 = \langle \gamma_i, \gamma_i \rangle = \frac{n}{2}(\theta_i^1 - \theta_i^2)' I(\theta_i)(\theta_i^1 - \theta_i^2),$$

and $\theta_i = \frac{\theta_i^1 + \theta_i^2}{2}$. Then, uniformly in $i$,

$$|\chi(u; m, \gamma^2) - g(u; \gamma_i^2)| \longrightarrow 0 \qquad (11)$$

as $n \to \infty$, $i = 1, \ldots, \kappa$.

Now, using the property of the non-central $\chi^2$ distribution (see [11], for instance) and Theorem 2, we can conclude that $\mathsf{E}\left(\frac{n\hat{\mathcal{J}}_i}{2}\right) = \gamma_i^2 + m + \mathcal{O}(n^{-3/2})$. Since the true value of the block discriminative power, $\frac{n\mathcal{J}_i}{2}$ is given by $\gamma_i^2$, in this context, the bias of each $\frac{n\hat{\mathcal{J}}_i}{2}$ is of order $m$, block size. This implies that when weighting by estimates, the true impact of each block of features is seriously overestimated. Moreover, the accumulation of the bias over the increasing number of blocks leads to the bias of the discriminant (8) of order $\mathcal{O}(\kappa/n)$. A reasonable goal, consistent with our approach would be to derive a down-weighting procedure, which can be provided by a properly chosen function $w$. As we will show in the present section this result can be obtained by minimizing the misclassification probability over all possible types of weighting.

We first extend the technique proposed in section 3 for assessing the classification accuracy of $\mathcal{D}_w(\mathbf{x}, \hat{\theta}^1, \hat{\theta}^2)$. This extension is fairly straight forward because the weighted discriminant also has an additive structure, which can be exploit in order to approximate the distribution of $\mathcal{D}_w(\mathbf{x}, \hat{\theta}^1, \hat{\theta}^2)$. Again evaluating asymptotic moments of the discriminant we obtain explicit expressions for the misclassification probabilities. The main diversity which has to be accounted for is that when using weighted discriminant, $w(\frac{n\hat{\mathcal{J}}_i(n)}{2})$ and $\mathcal{D}_i(\mathbf{x}_i, \hat{\theta}_i^1, \hat{\theta}_i^2)$ are statistically dependent being both constructed by the same set of data. In order to facilitate calculations of the asymptotic moments, we represent both $\mathcal{D}_i(\mathbf{x}_i, \hat{\theta}_i^1, \hat{\theta}_i^2)$ and $\frac{n\hat{\mathcal{J}}_i}{2}$ in terms of $T_i^1$ and $T_i^2$ introduced in section 3, which in turn makes it possible to find that

$$\mathsf{E}[\mathcal{D}_w(\mathbf{x}; \hat{\theta}^1, \hat{\theta}^2)] \to E(w), \quad \mathsf{Var}[\mathcal{D}_w(\mathbf{x}; \hat{\theta}^1, \hat{\theta}^2)] \to V(w),$$

as $n \to \infty$, where

$$E(w) = \rho \int \gamma^2 [\int w(u)\chi(u; m+2, \gamma^2) du] dH(\gamma^2),$$

$$V(w) = 2\rho \int [\int u w^2(u)\chi(u; m, \gamma^2) du] dH(\gamma^2). \quad (12)$$

Here the integration is performed over the non-central $\chi^2$ distribution represented by $\chi(u; m, \gamma^2)$ and over the limiting distribution $H(\gamma^2)$ of the discriminative power introduced in section 3. Observe that the former integration appears because, as we have proved in Theorem 2, $\frac{n\hat{\mathcal{J}}_i}{2}$ is approximately $\chi^2$ distributed. This assertion was used when performing the limiting step from summation over the number of blocks $\kappa$, to the integration over $\chi(u; m, \gamma^2)$. Since the third moment of $\mathcal{D}_w(\mathbf{x}, \hat{\theta}^1, \hat{\theta}^2)$ is of order $\mathcal{O}(\frac{1}{\sqrt{n}})$, we can make use of Gaussian approximation and get expressions for the misclassification probability.

**Theorem 3:** *Assuming that classification is performed by the weighted discriminant $\mathcal{D}_w(\mathbf{x}, \hat{\theta}^1, \hat{\theta}^2)$ where the weighting is governed by the factor $w(\frac{n\hat{\mathcal{J}}_i}{2})$, $i = 1, \ldots, \kappa$, the misclassification probabilities expressed by*

$$\mathcal{E}_1(w) = \Pr\left(\mathcal{D}_w(\mathbf{x}, \hat{\theta}^1, \hat{\theta}^2) \leq \pi_0 | \mathcal{C}(\mathbf{x}) = 1\right)$$

$$\mathcal{E}_2(w) = \Pr\left(\mathcal{D}_w(\mathbf{x}, \hat{\theta}^1, \hat{\theta}^2) > \pi_0 | \mathcal{C}(\mathbf{x}) = 2\right) \quad (13)$$

*where $\pi_0 = \ln \frac{\pi_2}{\pi_1}$, have the limits*

$$\mathcal{E}_1(w) \longrightarrow \Phi\left(-\frac{E(w) - \pi_0}{\sqrt{V(w)}}\right)$$

$$\mathcal{E}_2(w) \longrightarrow \Phi\left(-\frac{E(w) + \pi_0}{\sqrt{V(w)}}\right). \quad (14)$$

*as $n \to \infty$. Further, let $W$ be such class functions that for all $w(u) \in W$ both $E(w)$ and $V(w)$ do not equal to zero, then assuming that $\pi_1 = \pi_2$, and denoting $w_0(u) := \arg\min_{w(u) \in W} \mathcal{R}(w)$, we get*

$$w_0(u) = \frac{\int \gamma^2 \chi(u; m+2, \gamma^2) dH(\gamma^2)}{u \int \chi(u; m, \gamma^2) dH(\gamma^2)}. \quad (15)$$

*where $\mathcal{R}(w) = \frac{1}{2} \lim_{n \to \infty} (\mathcal{E}_1(w) + \mathcal{E}_2(w))$.*

It is not difficult to show that $w_0(u)$ is bounded, continuous for $u > 0$ and $w_0(u) \in W$. It is also straight forward to verify from (14) that the corresponding minimum value of the misclassification risk $\mathcal{R}$ is

$$\mathcal{R}(w_0) = \Phi\left(-\frac{1}{2}\sqrt{2\rho \int \frac{[\int \gamma^2 \chi(u; m+2, \gamma^2) dH(\gamma^2)]^2}{u \int \chi(u; m, \gamma^2) dH(\gamma^2)} du}\right). \quad (16)$$

The practical implementation of a weighting technique requires specification of the distribution $H(\gamma^2)$ of features discriminative power. To give an impression of how the weighting of the augmented discriminant by $w_0(u)$ works, we consider one simple choice of $H(\gamma^2)$.

**Example:** Distributions that can describe the *a priori* knowledge about the feature informativeness include e.g. a *point mass distribution*, $dH(\gamma^2) = 1$ concentrated in a certain point, $\gamma^2$. Using this type of distribution means that the inputs of all blocks into the distance $\mathcal{J}(\kappa)$ are identical so that the discriminative power of all blocks of features is assumed to be the same and equal to $\gamma^2$. Then in a view of (15) and given the point mass distribution $H(\gamma^2)$, the optimal weight function, $w_0$, turns out to be

$$w_0(u) = \frac{\gamma^2 \chi(u; m+2, \gamma^2)}{u\chi(u; m, \gamma^2)},$$

which according to (16) gives the limiting risk

$$\mathcal{R}(w_0) = \Phi\left(-\frac{1}{2}\sqrt{2\rho \int \frac{[\gamma^2\chi(u; m+2, \gamma^2)]^2}{u\chi(u; m, \gamma^2)} du}\right).$$

We now may understand the effect of weighting by comparing $\mathcal{R}(w_0)$ with $\mathcal{R}(1)$, i.e. with the misclassification risk when no weighting is involved and $w_0 = 1$. For this case, we use (12) as well as properties of the non-central $\chi^2$ distribution and find

$$E(1) = \rho \int \gamma^2 dH(\gamma^2) = \rho\gamma^2 \quad (17)$$

$$V(1) = \rho \int [\int u\chi(u; m, \gamma^2) du] dH(\gamma^2) = \rho(\gamma^2 + m),$$

which in turn gives

$$\mathcal{R}(1) = \Phi\left(-\frac{1}{2}\sqrt{2\rho\frac{\gamma^4}{\gamma^2 + m}}\right).$$

Using the standard arguments it is not difficult to show that

$$\int \frac{[\chi(u; m+2, \gamma^2)]^2}{u\chi(u; m, \gamma^2)} du \geq \frac{1}{\gamma^2 + m}.$$

Since $\Phi(y)$ is a decreasing function of $y$, we conclude that $\mathcal{R}(w_0) < \mathcal{R}(1)$.

Observe that the obtained result could be seen as somewhat counter intuitive: Assuming the true discriminative power to be equal for all blocks and thereby giving them equal weights, clearly should *not* effect classification accuracy. Our results however clearly indicate the decrease of misclassification risk when using weighting by $w_0$. A clue to the decrease of $\mathcal{R}(w_0)$ is provided by the results of Theorem 2, where we have shown that when using sample a based weighting technique, block discriminative power given by $\frac{n\tilde{J}}{2}$ is heavily overestimated in a high dimensional setting. The natural goal is hence to bring down the affect of the bias on the classification accuracy and in this case the results of Theorem 2 will likely be relevant.

The example presented here is an especially simple one meant to illustrate the principals involved; even in a specific case of a point mass distribution $H(\gamma^2)$, the optimal function $w_0$ established in Theorem 3 turns out to be sufficiently sensitive to the high dimensionality effects and provides the desirable down-weighting. As a further illustration of the weighting technique and the high dimensionality effect, we have in Figure 1 plotted $w_0$ and $\mathcal{R}(w_0)$ under different values of $m$, $\kappa$ and $\rho$. As expected each weight function places substantial part of its mass to the right tail (left panel) so that the block impacts with high deviations of estimates are down-weighted. Observe also that the weight function seems to be more "flat" as the block size $m$ increases. The right panel shows the asymptotic misclassification risk $\mathcal{R}(w_0)$ when weighting by $w_0$. Not surprising it is seen to be slowly decreasing given the smaller number of independent blocks in the network, i.e. when $\kappa = 4$ (dashed line) whereas embedding more independence in the network structures, i.e. letting $\kappa = 8$ and reducing the block size, lead to faster decreasing (solid line). Roughly speaking, given that the network structure is correct, the corresponding additive discriminant borrows strength from the block density which naturally results in a better classification accuracy. However, the design of these procedures should take into account discriminative power of the blocks *combined* with the effect of high dimensionality induced by the plug-in estimative approach.

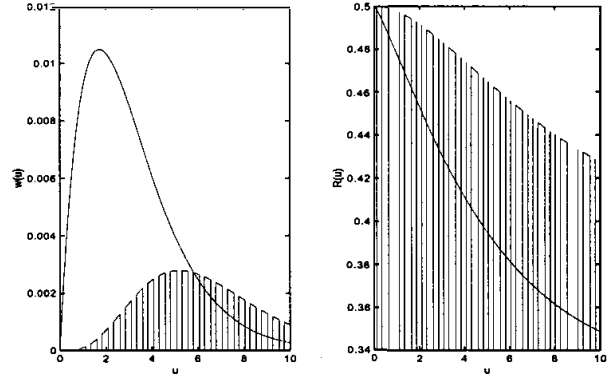

Figure 1: Optimal weight function $w_0(u)$ (left) and associated misclassification risk $\mathcal{R}(w_0)$ (right), given that $dH(\gamma^2) = 1$, $\gamma^2 = 1.8$ and $n = 36$. The behavior of $w_0(u)$ and $\mathcal{R}(w_0)$ with $m = 3$, $\kappa = 8$ and $\rho = 0.222$ (solid line); with $m = 6$, $\kappa = 4$ and $\rho = 0.111$ (dashed line).

## 5　Conclusions and scope for the future

We described the growing dimension asymptotic approach as a convenient tool for assessing classification accuracy of Bayesian network classifiers in a high dimensional setting. For a given network structure, the standard plug-in technique was used to estimate each of the class densities and then combine them into a discriminant function. The resulting procedure for both naive Bayes classifier and its augmented version pre-serves the attractive property - the discriminant has an additive form and as we have seen the Gauss approximation gives explicit expressions for misclassification probabilities in terms of the cross-entropy distance between the class densities and the high dimensionality factor.

In the process, we proposed a novel measure of the feature discriminative power which is given by the normalized cross-entropy distance between the classes. Furthermore, we introduced a weighting technique which makes it possible to take into account different discriminative powers of blocks of features within an augmented naive Bayes classifier. This technique being combined with the growing dimension asymptotic framework enables us to obtain an optimal, in a sense of asymptotic misclassification risk, type of weighting, given the network structure, size of the blocks and the distribution of their discriminative power.

The proposed technique can be extended in different directions. For example, to deal (given the blocks size) with several types of distributions of the discriminative power in order to select the form that "match" best the data structure and therefore is most useful for improving classification accuracy by weighting. Another direction may deal with feature selection: Observe that it is straight forward to extend the weighting technique to the selection technique by means of a discrete, (0/1) analog of weights, which reflect whether a block of features is included into classification. Due to the high dimensional consideration, incorporating a feature selection mechanism into the classifier may lead to better performance accuracy. Finally it is clear that both weighting and subset selection are applicable not only to classification, but also to parameter estimation and other tasks using Bayesian network models.

## Acknowledgments

Enlightening discussions with Per Uddholm are gratefully acknowledged. Tatjana Pavlenko was supported in part by Västernorrland County Council RDC under grant FoU:JA-2001-1001. Thanks to a nonymous reviewers for providing useful feedback about the presentation.